\documentclass[letterpaper, 10pt]{article} 
\usepackage[preprint]{tmlr}

\usepackage{amsmath,amsfonts,bm}

\def\eqref#1{equation~\ref{#1}}

\def\1{\bm{1}}

\DeclareMathAlphabet{\mathsfit}{\encodingdefault}{\sfdefault}{m}{sl}
\SetMathAlphabet{\mathsfit}{bold}{\encodingdefault}{\sfdefault}{bx}{n}

\usepackage{hyperref}
\usepackage{url}
\usepackage[pdftex]{graphicx}
\usepackage{caption}
\usepackage{subcaption}
\usepackage{float}
\usepackage{multirow}

\usepackage{times}
\usepackage{titlesec}

\titleformat{\section}
  {\bfseries\fontsize{10}{12}\selectfont\rmfamily} 
  {\thesection.}{0.5em}{} 

\titleformat{\subsection}
  {\bfseries\fontsize{10}{12}\selectfont\rmfamily}
  {\thesubsection}{0.5em}{}

\titleformat{\subsubsection}
  {\bfseries\fontsize{10}{12}\selectfont\rmfamily}
  {\thesubsubsection}{0.5em}{}

\newcommand{\webextramain}[1]{%
  \par\vspace{1em}%
  {\bfseries\fontsize{12}{14}\selectfont #1}\par\vspace{0.75em}%
}

\title{\vspace{-15pt}TabText: Language-Based Representations of Tabular Health Data for Predictive Modelling \vspace{-20pt}}

\author{\name Kimberly Villalobos Carballo \email kimberly.v@nyu.edu \\
      \addr Tandon School of Engineering \email +1 (617) 388 9911\\
      New York University \vspace{-5pt}
      \AND 
      \name Liangyuan Na \email  \\
      \addr Operations Research Center\\
      Massachusetts Institute of Technology \vspace{-5pt}
      \AND
      \name Yu Ma \email \\
      \addr Operations and Information Management\\
      University of Wisconsin, Madison \vspace{-5pt}
      \AND
      \name Léonard Boussioux \email  \\
      \addr Foster School of Business\\
      University of Washington, Foster School of Business \vspace{-5pt}
      \AND
      \name Cynthia Zeng \email \\
      \addr Stern School of Business\\
      New York University Abu Dhabi \vspace{-5pt}
      \AND
      \name Luis Soenksen \email \\
      \addr Abdul Latif Jameel Clinic for Machine Learning in Health\\
      Massachusetts Institute of Technology \vspace{-5pt}
      \AND
      \name Dimitris Bertsimas \email\\
      \addr Sloan School of Management\\
      Massachusetts Institute of Technology}

\def\month{MM}  
\def\year{YYYY} 
\def\openreview{\url{https://openreview.net/forum?id=XXXX}} 
\usepackage{array}
\newcolumntype{P}[1]{>{\centering\arraybackslash}p{#1}}
\usepackage{arydshln}

\usepackage{makecell}

\begin{document}

\maketitle
\vspace{-15pt}
\begin{abstract}

 \vspace{-5pt}\textbf{Background:} Tabular medical records remain the most readily available data format for applying machine learning in healthcare. However, traditional data preprocessing ignores valuable contextual information in tables and requires substantial manual cleaning and harmonisation, creating a bottleneck for model development. \textbf{Methods:} We introduce TabText, a preprocessing and feature extraction method that leverages contextual information and streamlines the curation of tabular medical data. This method converts tables into contextual language and applies pretrained large language models (LLMs) to generate task-independent numerical representations. These fixed embeddings are then used as input for various predictive tasks. TabText was evaluated on nine inpatient flow prediction tasks (e.g., ICU admission, discharge, mortality) using electronic medical records across six hospitals from a US health system,
 and on nine publicly available datasets from the UCI Machine Learning Repository, covering tasks such as cancer diagnosis, recurrence, and survival. \textbf{Findings:}
TabText models trained on unprocessed data from a single hospital (572,964 patient-days, Jan 2018–Dec 2020) achieved accurate performance (AUC 0·75–0·94) when tested prospectively on 265,917 patient-days from Jan 2021–Apr 2022, and generalised well to five additional hospitals not used for training. When augmenting preprocessed tabular records with these contextual embeddings, out-of-sample AUC improved by up to 4 additive percentage points in challenging tasks such as ICU transfer and breast cancer recurrence, while providing little to no benefit for already high-performing tasks. Findings were consistent across both private and public datasets. \textbf{Interpretation:} Converting tables into language allows to generate task‑independent embeddings that retain contextual information. These embeddings enable the rapid development of high-performing predictive models that can be applied across institutions even when data formats differ. This approach can accelerate the development of predictive analytics in healthcare and enhance performance in difficult tasks, such as those with limited data. \textbf{Funding:} No funding was received for this work.
\end{abstract}

\section{Introduction}\label{sec:introduction}
Tabular datasets are one of the most common formats for storing and analysing health-related information. They span a wide range of contexts, from electronic medical records in hospitals to disease registries, public health records, and clinical trial data. These datasets often combine numerical measurements, categorical variables, and derived indicators to represent individuals, samples, or events.~\citep{rajkomar2019machine} However, they are rarely standardised: formats, coding schemes, and variable definitions can vary greatly between sources, and missing or inconsistently recorded values are common.~\citep{li2024scoping} Preparing such data for machine learning typically requires substantial manual effort, creating a bottleneck for timely model development.~\citep{rajkomar2018scalable} Moreover, conventional processing approaches focus only on the recorded values and often ignore accompanying contextual information, such as variable names, measurement units, or table-level metadata, which could provide valuable signals to improve predictive performance.~\citep{lee2025clinical}

Language, in contrast to rigid tabular structures, offers a flexible modality for representing heterogeneous and non-standardised health information, enabling to capture both the recorded values and their context. In the health domain, efforts to reduce manual processing and improve the predictive utility of tabular data via language have followed three main directions. The first is the development of domain-specific foundation models—such as ClinicalBERT,~\citep{clinicalBERT} BioGPT,~\citep{Luo_2022} and Clinical-Longformer~\citep{li2022clinicallongformer}—pretrained on large corpora of biomedical or clinical text. These models require text inputs and do not natively process structured tables, but converting tables into language allows the prediction problem to be reformulated as a natural language task.~\citep{pmlr-v206-hegselmann23a, wang2023meditab} The second approach develops foundation models trained from scratch to operate directly on sequences of structured elements within electronic health records, as in BEHRT~\citep{li2020behrt} and related architectures. A third, more recent direction converts tables into text, leverages pretrained language models to produce contextual embeddings for various data streams, and combines them with additional architectures—such as attention or linear layers.~\citep{lee2025clinical, yang2025leveraging}

Previous studies in this third line of work have shown that contextual embeddings derived from tables can outperform tabular models—such as trees, linear models, and transformers—as well as direct prompting approaches. However, previous works generate embeddings for each data stream separately (e.g., vitals, labs, medications), and further training of fixed neural network architectures is required to learn how to combine them into a single representation that depends on the specific prediction task. A simple framework for constructing contextual embeddings as general-purpose representations of tabular data—applicable across different models and tasks—has not yet been established. Moreover, prior work has not investigated the effect of augmenting tabular values with contextual embeddings, and no embedding-based approach has demonstrated strong predictive performance across institutions.

Motivated by this gap, we introduce TabText, a flexible preprocessing framework that leverages language to convert tabular health records into fixed embeddings that can be reused across multiple prediction tasks and machine learning models. TabText reduces the need for manual data preprocessing and streamlines the preparation of heterogeneous health datasets for predictive modelling. We evaluate its predictive performance across six hospitals and examine the effect of augmenting tabular data with contextual embeddings in both inpatient electronic health records (EHR) and publicly available health datasets. Our results show that with only minimal preprocessing, TabText enables the development of predictive models that achieve strong out-of-sample performance and generalise well across external datasets. We further demonstrate that augmenting the tabular feature space with TabText embeddings substantially improves performance in difficult tasks such as ICU transfers and breast cancer recurrence, while providing little to no benefit for high-performing tasks.
\begin{figure}[ht]
    \centering
    \includegraphics[width=0.9\linewidth]{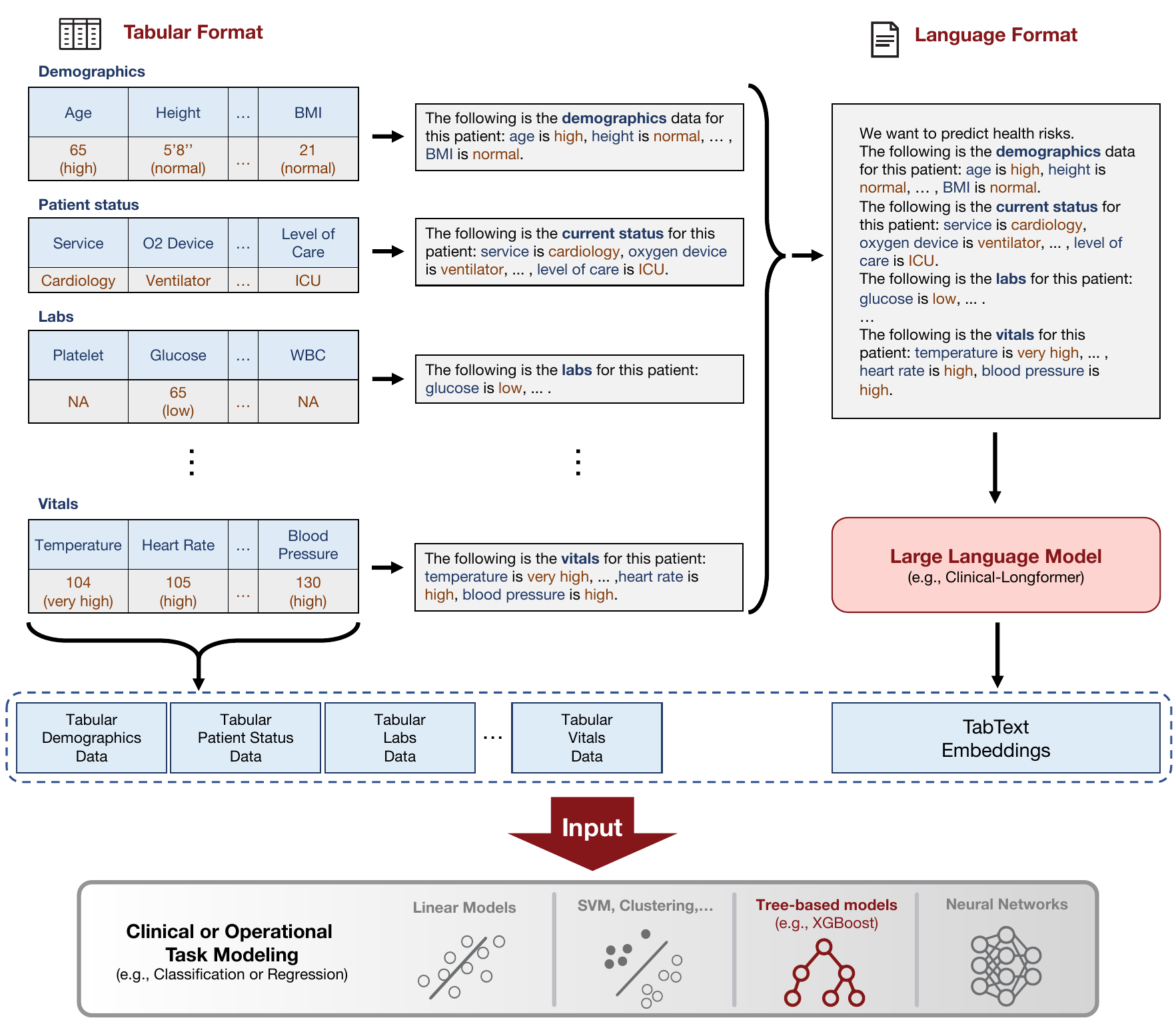}
    \caption{End-to-end TabText framework integrating diverse tabular data sources into a language format that can be combined into a single patient description paragraph. This language representation is then fed to existing pretrained LLMs to generate embeddings for downstream prediction tasks.}
    \label{fig:diagram}
\end{figure}

\section{Methods} \label{sec:Methods}
TabText transforms each table into descriptive sentences that are then concatenated into a single summary paragraph for each data subject. This text contains the attributes of the columns with their corresponding values and potentially other available contextual information. A pretrained LLM subsequently processes this language-based representation to generate fixed-size embeddings that encode the textual data. Finally, these embeddings are used to train any standard machine learning model for downstream prediction tasks. Figure \ref{fig:diagram} illustrates the overall TabText framework. 

In our methodology we seek to identify a pretrained language model, a systematic strategy for sentence construction, and a training approach that together generalise well across predictive tasks. Specifically, we aim to generate a single TabText embedding per patient that performs well across multiple downstream predictions rather than tailoring embeddings to individual tasks. This approach not only promotes representations that capture all the relevant information in the input text, but also reduces computational overhead by eliminating the need to recompute embeddings for new tasks.

To this end, we consider nine different binary classification tasks: whether patients are discharged or not within the next 24 hours (resp. 48 hours); whether the patient will enter (resp. leave) the intensive care unit (ICU) for patients currently not in the ICU (resp. in the ICU) within the next 24 hours (resp. 48 hours); whether each patient will die in the next 24 hours (resp. 48 hours); and whether patients die or not at the end of their hospital stay. We utilise a private real-world dataset comprised of 63 columns corresponding to different laboratory results of inpatients in a teaching hospital in 2017. We use $60,000$ data samples for training and validation, and $10,000$ for testing, where each data point corresponds to a patient day. All analyses in Sections \ref{sec:LLM}, \ref{sec:language}, and \ref{sec:finetune} are based on this dataset and the nine prediction tasks described above, which we use to develop and refine the TabText framework.

\begin{figure}[ht]
    \centering
    \begin{subfigure}[b]{\linewidth}
    \includegraphics[width=\textwidth]{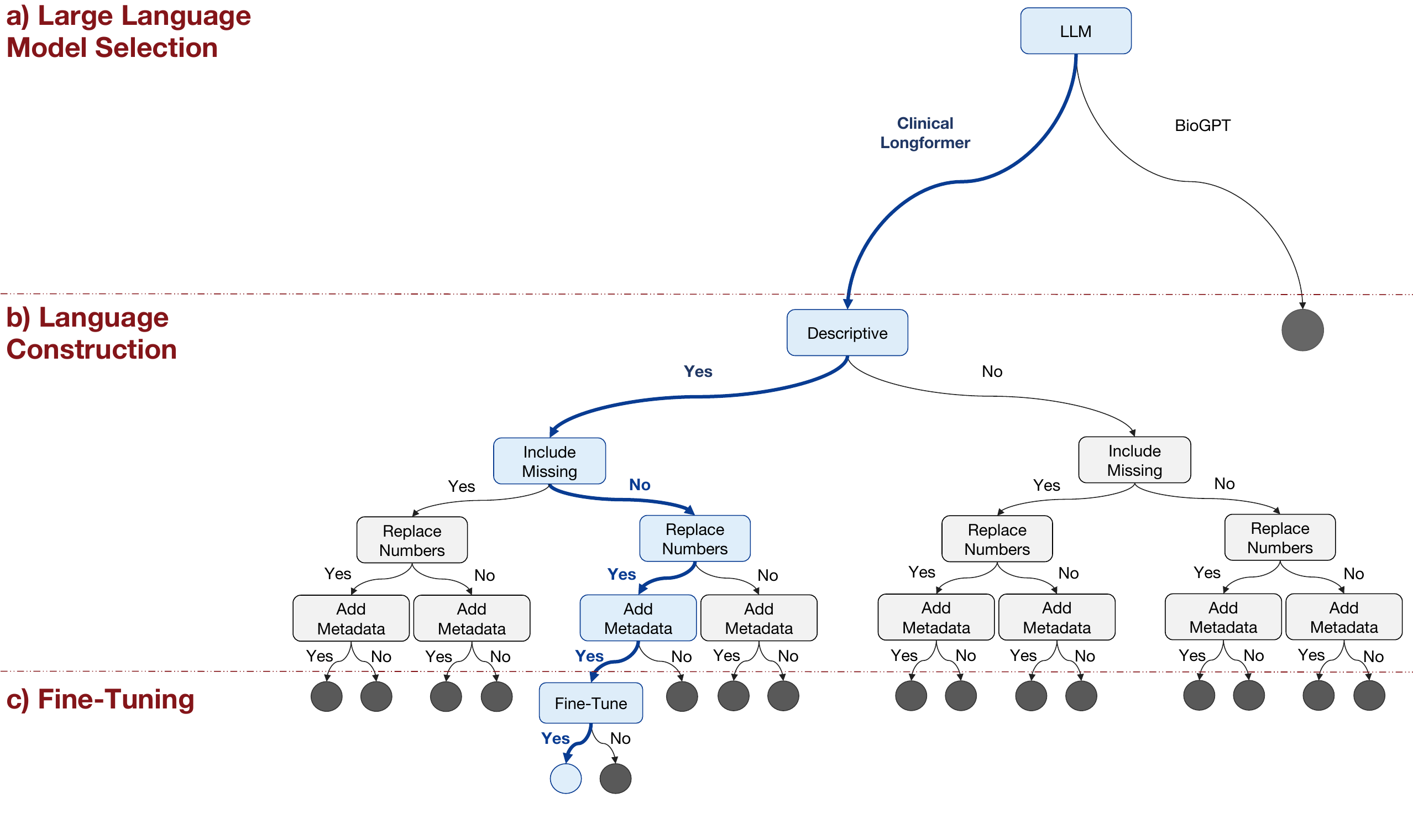}
    \end{subfigure}\\
    \begin{subfigure}[b]{0.32\linewidth}
    \includegraphics[width=\linewidth]{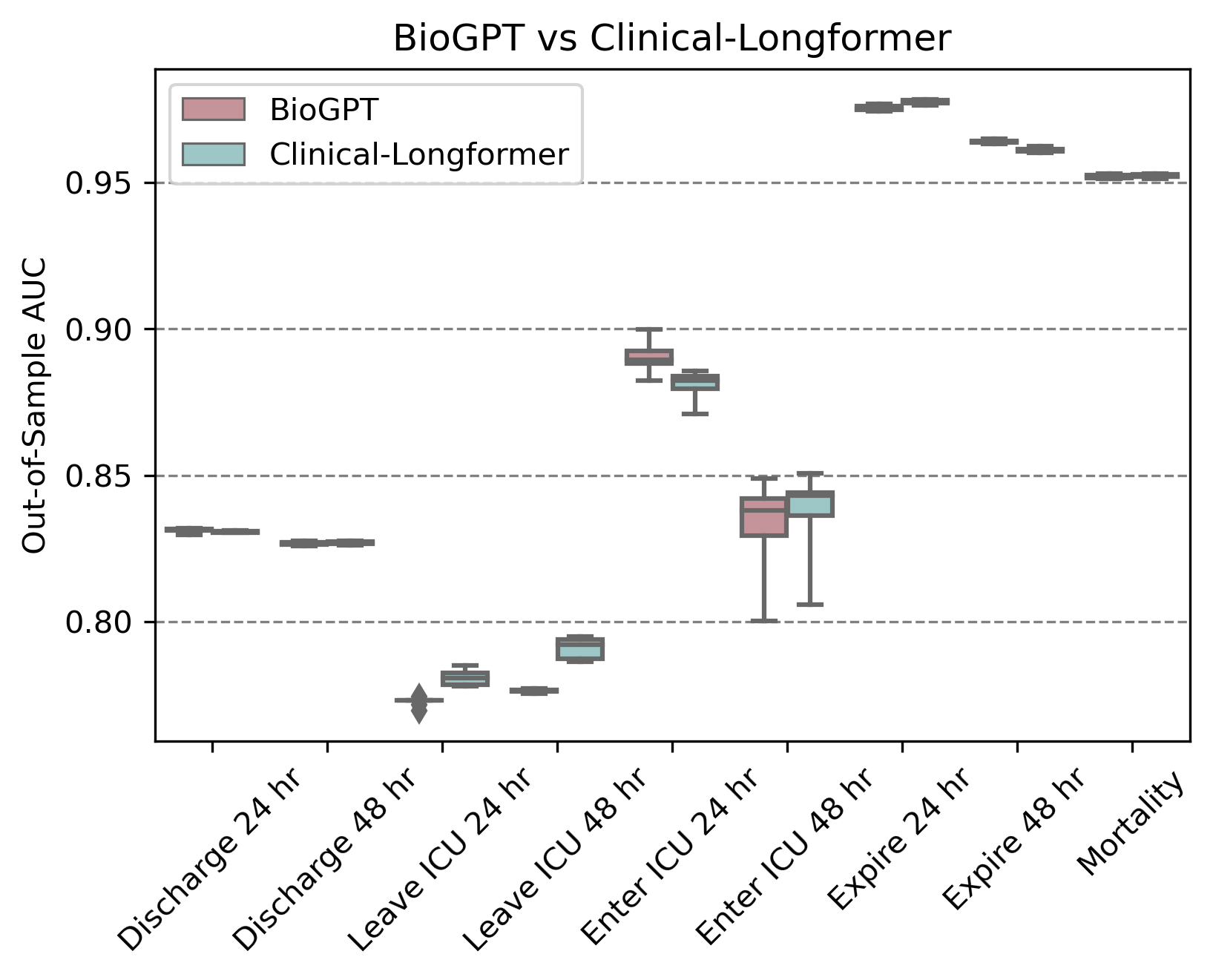}
    \caption{LLM Selection.}
    \label{LLM}
    \end{subfigure}
    \hfill
    \begin{subfigure}[b]{0.32\linewidth}
    \includegraphics[width=\linewidth]{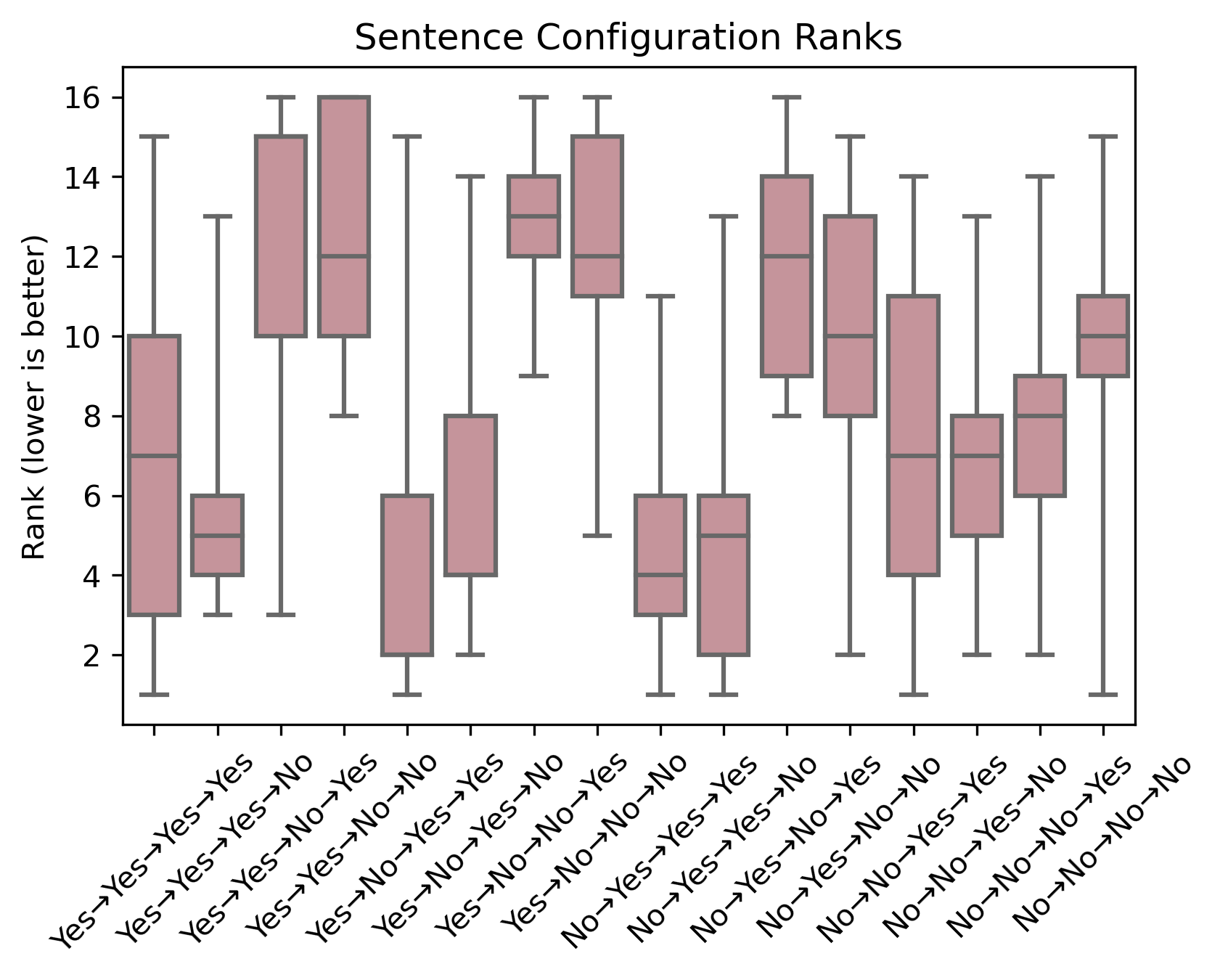}
    \caption{Language Construction.}
    \label{sentence-construction}
    \end{subfigure}
    \hfill
    \begin{subfigure}[b]{0.32\linewidth}
    \includegraphics[width=\linewidth]{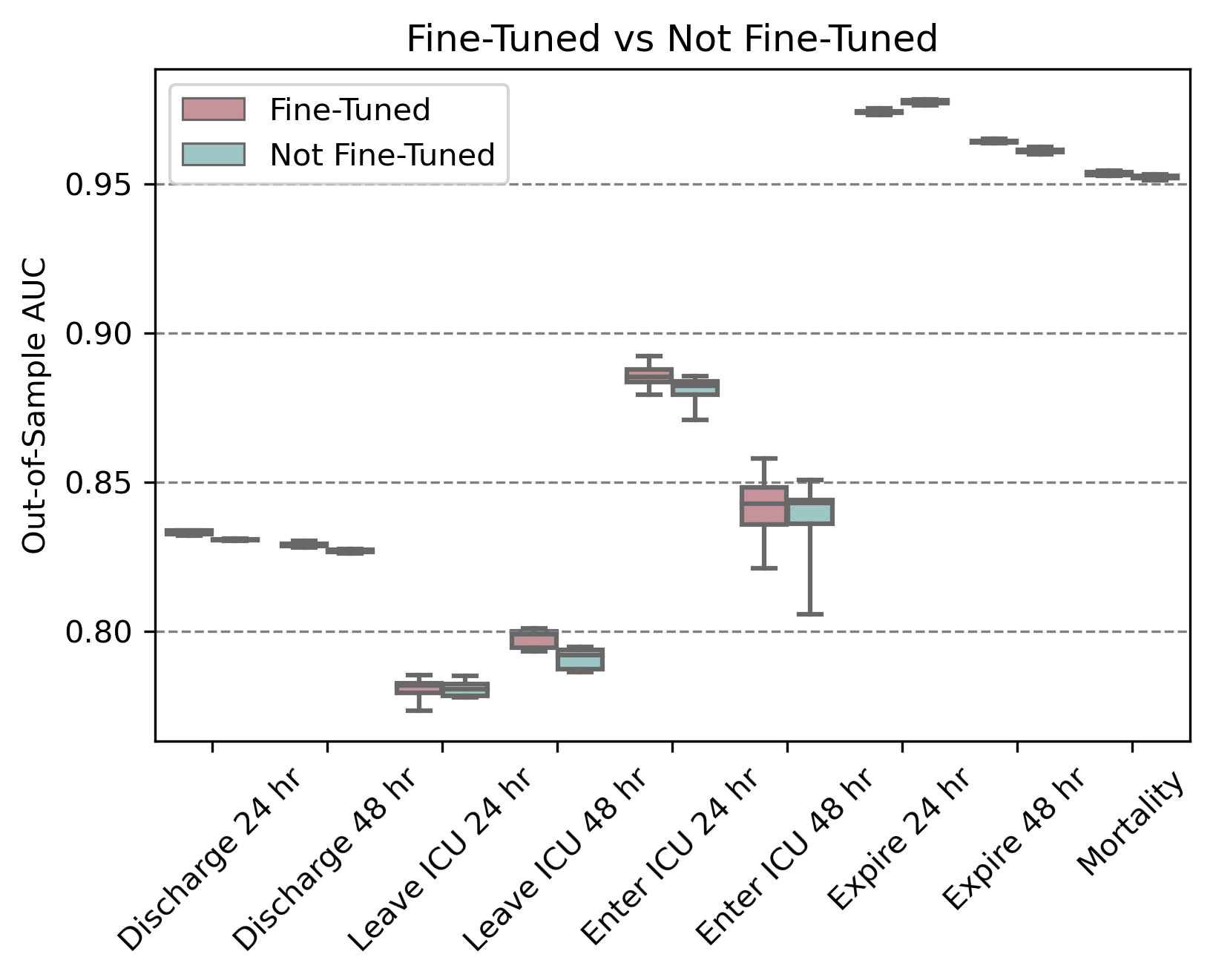}
    \caption{Self-Supervised Fine-Tuning.}
    \label{fine-tune}
    \end{subfigure}
\caption{Overview of our overall methodology. We begin by selecting an LLM. Figure (a) shows that BioGPT and Clinical Longformer achieve similar results, and we therefore chose Clinical Longformer, as it allows for input texts of larger sizes. Then, we look for the best language representations of the original patient data. In Figure (b) we observe the boxplots of the ranks for different sentence configurations, which were tested across different prediction tasks (lower rank is better). We select the configuration with the lowest median ranking; specifically, we use descriptive language, omit missing values, replace numerical values with text, and include metadata. Lastly, we use word-masking to fine-tune the LLM with the chosen sentence configuration, as it leads to better performance (Figure (c)). }\label{fig:meth_diagram}
\end{figure}

\subsection{LLM Selection}\label{sec:LLM}
We consider two transformer-based language models: BioGPT,~\citep{Luo_2022} pretrained on 15 million PubMed abstracts, and Clinical-Longformer, \citep{li2022clinicallongformer} which was initialised from Longformer and further pre‑trained on clinical notes from the Medical Information Mart for Intensive Care III (MIMIC-III) database. \citep{johnson2016mimic} 
These models differ in their training objectives: BioGPT follows a causal, next-token prediction approach, whereas Clinical-Longformer uses a masked language modelling strategy. Evaluating both allows us to investigate whether the underlying pretraining methodology influences how effectively language representations of tabular data support downstream predictive tasks. Following the TabText framework, we convert the tabular data into simple text: for each row, the cell from column ``attribute'' with value X is transformed into ``attribute: X'', and the texts from all columns are concatenated into a single sentence with the comma character. We next create TabText embeddings and finally use them as input to train gradient-boosted tree models for the nine tasks of interest. Figure \ref{LLM} shows the boxplots for the out-of-sample area under the ROC curve (AUC) over 10 random 75\%-25\% train-validation splits for each task and each model. There were no statistically significant differences in performance between the models, and we therefore chose the Clinical-Longformer model, as it allows for input text of larger size.

\subsection{Language Construction}\label{sec:language}
The versatility of language creates a challenge for consistency, as multiple textual expressions can convey the same information. The TabText framework creates a single paragraph for each data sample (e.g., for each patient day) as follows: we first create a sentence for each column in each table. Next, for each table, we concatenate contextual information and the sentences of its columns using the colon (``:'') and comma (``,'') characters, respectively. We then merge the text from all tables into a single paragraph using the period (``.'') character. While the exact punctuation doesn't significantly impact Clinical-Longformer, \citep{punctuation} the exact text chosen to build each sentence might have a larger impact on the final embedding. 

We investigate several strategies for sentence construction, including whether to use descriptive phrasing (e.g., “attribute is X” versus “attribute: X”), how to handle missing values (explicitly noting “attribute is missing” versus omitting the column), whether to discretise continuous variables into categorical text bins (e.g., mapping numerical values to “very low,” “low,” “normal,” “high,” or “very high”), and whether to incorporate metadata such as table descriptions; exact implementation details are provided in Appendix \ref{sec:language_construction}.

For each of the 16 possible sentence configurations, we use default values of the Clinical-Longformer model to obtain TabText embeddings that are given as input to a gradient-boosted tree model. In Figure \ref{sentence-construction} we show the boxplot for the rank achieved by each configuration across tasks, where lower numbers are better. We choose the sentence configuration with the lowest median ranking; specifically, we use descriptive language, omit missing values from the text, replace numerical values with text, and include metadata.

\subsection{Self-Supervised Fine-Tuning}\label{sec:finetune}
Although Clinical-Longformer was pretrained on large language datasets, we investigated whether additional masked-language pretraining on the constructed text could further improve performance. Specifically, we convert our training data into language following the sentence configuration selected in Section \ref{sec:language}, and we use it to fine-tune Clinical-Longformer following its original training methodology, which consists of self-supervised masked word prediction. We then generate embeddings that are given as input to a gradient-boosted tree model. We show in Figure \ref{fig:meth_diagram} the boxplots for the out-of-sample AUC over 10 random 75\%-25\% train-validation splits for each task. We observed that fine-tuning the model with our tabular-based language data yields statistically significant (p <0·0001) improvements for eight out of the nine classification tasks, and we therefore include word-masking fine-tuning prior to extracting embeddings.

\section{Results}\label{sec:results}
We evaluate TabText on nine inpatient flow prediction tasks using EHR from six hospitals, and on nine publicly available healthcare datasets covering diagnosis, recurrence, and survival. Our goal is to assess generalisation across hospitals with minimal preprocessing and to quantify the added value of augmenting structured features with TabText embeddings. For all experiments, we converted tables to text using the best-performing sentence configuration (skipping sentences for missing values, replacing numbers with text, using descriptive language, and adding metadata) and fine-tuned Clinical-Longformer on randomly sampled training examples (from the training set only, ensuring no overlap with validation or test data) via word-masking with default hyperparameters. We then extracted 768-dimensional embeddings by mean-pooling the last hidden layer. The best models were chosen based on performance on the validation set. Additional implementation details can be found in Appendix \ref{sec:experiment_details}. 

\subsection{Private Hospital Datasets: Patient Flow Predictions}\label{subsec:PatFlow}
We apply the TabText framework on the same nine patient flow predictive tasks described in Section \ref{sec:Methods}, but this time using a much larger dataset that contains medical records of all inpatients over a four-year period from a private hospital that we call HA. Each data point represents a patient day over the period January 2018 - April 2022. There are 160 columns of different patient attributes on demographics, patient status, vital signs, laboratory results, diagnoses, treatments, and other information. We emphasise that although the dataset used for the development of the TabText methodology in Section \ref{sec:Methods} was also obtained from hospital HA, it only contains non-overlapping data from 2017. The summary of the available dataset and data sizes utilised can be found in Tables \ref{table:data_sizes} and \ref{tab:summary_tables} in Appendix \ref{sec:datasets_info}.

\subsubsection{Strong Out-of-Sample and Cross-Hospital Performance with Minimal Preprocessing}
\label{subsec:rawdata}
We first leverage the TabText framework to replace the heavy lifting of data cleaning by simply creating a text representation for each data sample using the information as it appears in the raw data tables that we received from the hospital. In particular, columns that require data cleaning to convert to appropriate data types can instead be treated as free text. For example, the sentence corresponding to a column for a sedation score with the value ``-4 $\to$ deep sedation'' can be considered as a categorical column and written as ``sedation score is -4 $\to$ deep sedation'', as opposed to parsing the original string into a numeric value of -4 as part of the traditional preprocessing steps.

\begin{table}[ht]
\centering
\begin{tabular}{lcccccc}
\hline
\hline
Task & HA & HB & HC & HD & HE & HF \\
\hline
Discharge 24 hr  & 0·792 & 0·759 & 0·758 & 0·775 & 0·772 & 0·751 \\
Discharge 48 hr  & 0·778 & 0·746 & 0·740 & 0·765 & 0·760 & 0·744 \\
Leave ICU 24 hr  & 0·836 & 0·843 & 0·797 & 0·840 & 0·866 & 0·802 \\
Leave ICU 48 hr  & 0·830 & 0·826 & 0·787 & 0·829 & 0·847 & 0·795 \\
Enter ICU 24 hr  & 0·794 & 0·772 & 0·750 & 0·781 & 0·810 & 0·795 \\
Enter ICU 48 hr  & 0·745 & 0·719 & 0·694 & 0·700 & 0·758 & 0·746 \\
Expire 24 hr     & 0·941 & 0·927 & 0·913 & 0·924 & 0·922 & 0·912 \\
Expire 48 hr     & 0·927 & 0·916 & 0·892 & 0·904 & 0·910 & 0·896 \\
Mortality      & 0·884 & 0·890 & 0·865 & 0·888 & 0·900 & 0·857 \\
\hline
\hline
\end{tabular}
\caption{Out-of-sample AUCs for hospitals HA–HF over the period January 2021 – April 2022. Models were trained on HA data from 2018 to 2020 with a temporally separated train–validation–test split, reflecting real-world deployment conditions. }\label{tab:not-processed}
\end{table}

Only minimal manual work was required, limited to constructing meta-information about the tables and columns. For model development, we adopt a temporal train–validation–test split to closely reflect real-world deployment conditions. Specifically, models were trained on data from hospital HA for the period January 2018 to February 2020, hyperparameters were selected using a validation period from March 2020 to December 2020, and final evaluation was conducted both out-of-sample on HA and on other hospitals (HB–HF) for the period January 2021 to April 2022. This design avoids information leakage across time and simulates real-world deployment, where models must generalise to future patients; temporal separation is especially important given the major changes in patient populations and clinical practices during the COVID-19 pandemic. For the downstream prediction models, we tried both gradient-boosted trees and linear models, selecting the one with the best performance for each task. These hospitals are part of the same health system but differ substantially in patient populations, geographic location, and services offered. Moreover, the electronic health record data are not standardised across sites, with inconsistencies in how values are recorded and formatted. More information about these hospitals is provided in Table \ref{all_hosp} in Appendix \ref{sec:datasets_info}.

As shown in Table \ref{tab:not-processed}, the baseline TabText models with minimally processed data already achieve high AUCs across all institutions, reaching practically implementable benchmarks for the hospital system. Importantly, the models trained on HA generalise well to other hospitals, indicating that the TabText embeddings effectively capture patient information that is broadly relevant. In some cases—such as Leave ICU 24 hr—the performance at other hospitals (HB, HD, HE) even exceeds that observed at HA. We observe lowest performance for the Enter ICU 48 hr task, which is a notoriously difficult classification task. \citep{na2023hartford}

\subsubsection{Enhanced Performance on Difficult Tasks via TabText Embeddings}\label{subsec:Context}
We investigate whether augmenting tabular data with context via TabText embeddings provides additional benefit compared with using the tables alone. To this end, we perform feature processing following Na et al. \cite{na2023hartford}, which includes merging raw tables, parsing string fields, encoding categorical variables, constructing derived features, and imputing missing values, as described in Appendix \ref{sec:data_cleaning}. 
We next group all features into 6 tables (see Table~\ref{tab:summary_tables}), which we combine and feed into the TabText Framework from Figure~\ref{fig:diagram} to obtain TabText embeddings.
\begin{figure}[htp]
    \centering
    \includegraphics[width=\linewidth]{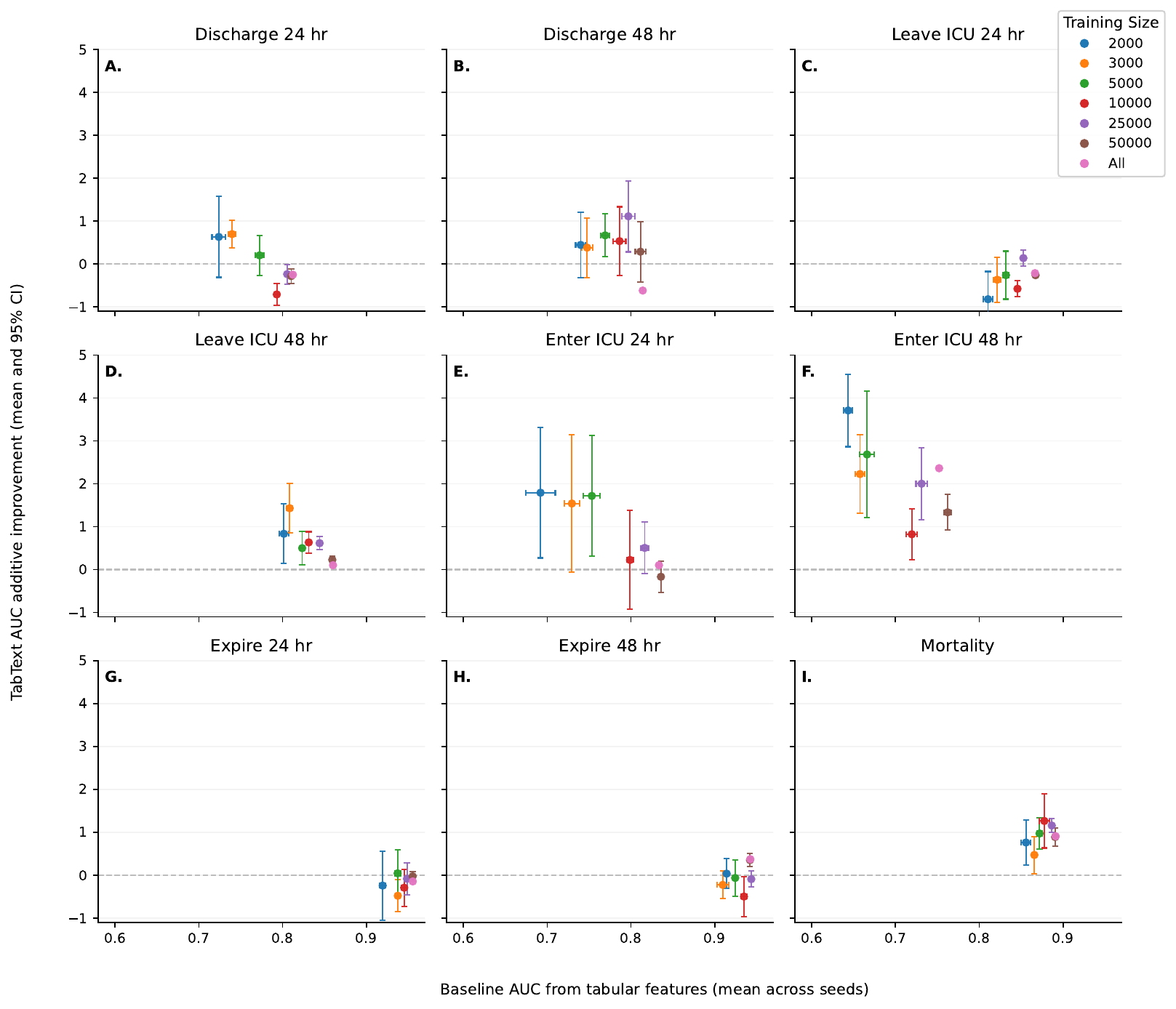}
\caption{Each panel shows TabText’s additive AUC improvement vs. the mean AUC obtained using the tabular features only; markers encode training size and vertical bars correspond to 95\% CIs. Improvements are largest for harder tasks and low-data regimes, and diminish as sample size and baseline AUC increase.}
    \label{tab:auc_processed}
\end{figure}

Unlike the previous section—where a fixed validation split was chosen to approximate deployment conditions and highlight the best achievable performance in practice—our focus here is on analysing potential gains from augmenting tables with TabText embeddings at different training sizes. To enable direct comparison with the previous results, we used the same temporal design as in Section \ref{subsec:rawdata}, holding out data from 2021 to April 2022 as a fixed test set, but varying the size of the training set. Within the training window (2018–2020), we evaluate TabText at multiple training sizes (2000; 3000; 5000; 10,000; 25,000; 50,000, and all available patient-days; see Table \ref{table:data_sizes}) with 10 random 75\%–25\% temporal train-validation splits for each experiment. Gradient-boosted trees were consistently selected as the best-performing model across tasks. For each task and training size, we report the mean test AUC using tabular features only (Tab AUC) on the x-axis, and the additive improvement in percentage points from augmenting features with TabText embeddings (TabText AUC - Tab AUC) on the y-axis. For the latter we show average and 95\% confidence intervals (CIs) computed as two-sided Student’s t intervals across the 10 seeds with 9 degrees of freedom.

As shown in Figure \ref{tab:auc_processed}, augmenting tabular features with TabText yields the largest improvements when the tabular-only baseline model is weaker (AUC $\leq $0·80) and training data are limited (training sizes $\leq 10k$), typically adding $~1$ to $3$ percentage points with confidence intervals mostly above the zero line. As sample size and baseline AUC increase, improvements shrink. Improvements are most pronounced for ICU-entry tasks, modest gains for discharge and ICU-leave, and negligible or slightly negative effects for short-horizon mortality/expiration, where tabular models already achieve AUC $\geq $0·93.

\subsection{\textcolor{black}{Public Datasets}}\label{subsec:PublicDatasets}
{\color{black}
To examine performance beyond EHR data, we also evaluated TabText on nine public healthcare datasets from the UCI Machine Learning Repository. \citep{kelly2023uci} Additional metadata information used for each dataset is shown in Tables \ref{tab:external_datasets} and \ref{tab:datasets-info} in Appendix \ref{sec:data_cleaning}. We compare three types of classifiers—FT-Transformer (FT),~\citep{gorishniy2021revisiting} Logistic Regression models (LR),~\citep{ridge,lasso,elasticnet} and XGBoost (XGB)~\citep{chen2015xgboost}—under three different input settings: tabular features only, text features only via TabText embeddings, and the concatenation of both tabular and text embeddings. FT-Transformer with TabText embeddings was excluded due to high computational cost and low preliminary performance. We also note that FT-Transformer applies standardisation and quantile-based transformations during preprocessing, which provides a natural point of comparison with the discretisation steps used to obtain TabText embeddings.

\begin{table}[htp]
\centering
\color{black}
\begin{tabular}{P{1.4cm}|*{3}{P{1.4cm}}|*{2}{P{1.4cm}}|*{2}{P{1.4cm}}|P{1.4cm}}
\hline\hline
\multirow{3}{*}{\textbf{Dataset}} & \multicolumn{3}{c|}{\textbf{Tabular Approach}} & \multicolumn{4}{c|}{\textbf{TabText Approach}} & \multirow{3}{*}{\textbf{$\Delta$ AUC}} \\
\cline{2-8}
& \multicolumn{3}{c|}{\textbf{Tabular}} & \multicolumn{2}{c|}{\textbf{Tabular + Text}} & \multicolumn{2}{c|}{\textbf{Text}} & \\
\cdashline{2-8}
  & \textbf{FT} & \textbf{LR} & \textbf{XGB} & \textbf{LR} & \textbf{XGB} & \textbf{LR} & \textbf{XGB} & \\
\hline
BC  & 0·703 & 0·716 & 0·695 & \textbf{0·755} & 0·733 & 0·753 & 0·728 & \textcolor{green!60!black}{+0·039} \\
    & (0·033) & (0·058) & (0·052) & (0·032) & (0·028) & (0·030) & (0·041) & \\[5pt]
HD  & \textbf{0·801} & 0·732 & 0·737 & 0·736 & 0·766 & 0·791 & 0·758 & \textcolor{red!70!black}{-0·010} \\
    & (0·020) & (0·056) & (0·029) & (0·058) & (0·023) & (0·019) & (0·025) & \\[5pt]
TS  & 0·565 & 0·603 & 0·589 & \textbf{0·612} & 0·606 & 0·609 & 0·604 & \textcolor{green!60!black}{+0·009} \\
    & (0·063) & (0·066) & (0·070) & (0·075) & (0·058) & (0·046) & (0·055) & \\[5pt]
AS  & 0·723 & 0·631 & 0·712 & 0·688 & 0·714 & \textbf{0·739} & 0·723 & \textcolor{green!60!black}{+0·016} \\
    & (0·029) & (0·056) & (0·028) & (0·047) & (0·049) & (0·035) & (0·034) & \\[5pt]
CP  & 0·768 & 0·771 & {0·775} & 0·750 & \textbf{0·780} & 0·670 & 0·660 & \textcolor{green!60!black}{+0·005} \\
    & (0·050) & (0·052) & (0·042) & (0·054) & (0·043) & (0·033) & (0·027) & \\[5pt]
CC  & 0·583 & 0·614 & {0·648} & 0·672 & 0·650 & \textbf{0·676} & 0·634 & \textcolor{green!60!black}{+0·028} \\
    & (0·049) & (0·033) & (0·045) & (0·044) & (0·046) & (0·050) & (0·041) & \\[5pt]
DR  & 0·976 & 0·968 & {0·990} & 0·974 & \textbf{0·993} & 0·974 & 0·990 & \textcolor{green!60!black}{+0·003} \\
    & (0·008) & (0·010) & (0·005) & (0·009) & (0·004) & (0·009) & (0·008) & \\[5pt]
MM  & 0·892 & 0·879 & {0·894} & \textbf{0·897} & 0·880 & 0·893 & 0·878 & \textcolor{green!60!black}{+0·003} \\
    & (0·013) & (0·014) & (0·012) & (0·010) & (0·016) & (0·011) & (0·012) & \\[5pt]
HS  & 0·634 & {0·667} &{0·667} & \textbf{0·681} & 0·624 & 0·643 & 0·632 & \textcolor{green!60!black}{+0·014} \\
    & (0·075) & (0·051) & (0·048) & (0·039) & (0·049) & (0·041) & (0·028) & \\
\hline\hline
\end{tabular}
\caption{\textcolor{black}{Average out-of-sample AUCs across 10 random data splits (with 95\% confidence intervals in parenthesis). Highest mean per dataset is bolded. The last column shows additive improvement in AUC of the best TabText model over the best Tabular model. Positive improvements in green, negative in red.}}
\label{tab:public-datasets}
\end{table}

Tables \ref{tab:public-datasets} presents the results across 10 random 60\%–20\%–20\% train–validation–test splits, including mean and 95\% CIs computed as two-sided Student’s t intervals across splits with 9 degrees of freedom. The first group of columns corresponds to models trained using only tabular features. Among these, FT-Transformer and XGBoost each obtain the highest AUCs on multiple datasets, with no single model dominating across all tasks. Logistic regression generally underperforms compared to the other two tabular models. The second group of columns reports results from TabText models, which use the text-derived embeddings. In most datasets, at least one TabText variant achieves a higher AUC than the best tabular-only model. For example, in the average-case analysis, TabText provides notable gains for BC (+3.9\%), AS (+1.6\%), and CC (+2.8\%), and 
only one dataset (HD) shows lower average performance for TabText (–1\%).
The largest gains occur in datasets for which the Tabular approach achieves low performance (BC, TS, AS, HS), consistent with our previous findings.

\section{Discussion}

We developed TabText, a framework that converts tabular healthcare data into descriptive text and generates task-independent embeddings using pretrained language models. Across inpatient flow prediction tasks at a large hospital system and multiple public health datasets, TabText enabled high-performing models from minimally processed data and generalised well across hospitals with different populations and data formats. Furthermore, augmenting tabular features with contextual information provided measurable gains for difficult predictions such as ICU transfer and breast cancer recurrence.

These findings complement and extend previous work on language-based representations of structured health data. Earlier studies have shown that representing tabular health data as language and extracting embeddings can improve predictive performance. However; previous works largely relied on separate embeddings for each table as sub-components of larger neural network architectures. A general approach for creating contextual embeddings of tabular data that can serve as task-agnostic representations across different models and prediction settings remained lacking. In addition, the potential benefits of enriching tabular values with contextual information was largely unexplored, and existing embedding-based methods had not demonstrated strong predictive performance across external datasets.

TabText offers important advantages. First, the framework reduces the manual effort required to clean and harmonise data, producing deployable baselines within hours of receiving raw tables. Second, its cross-hospital performance indicates that contextual information encoded in the embeddings captures broadly relevant patient features, even across sites with different patient populations and data formats. Third, the framework is flexible: it can be paired with standard classifiers such as gradient-boosted trees, logistic regression, or transformer architectures.

There are also important limitations. Embedding extraction introduces computational overhead compared with conventional tabular pipelines, although in our experiments, runtimes remained practical as the patient embedding only needs to be computed once and can then be leveraged across tasks and model types. The embeddings themselves are less interpretable than native tabular features, and additional tools are required to map predictions back to meaningful clinical variables. Our evaluations focused primarily on one health system; although cross-hospital performance was strong, broader validation across other networks, care settings, and institutions is needed. Finally, as with all machine learning approaches, errors in the underlying data may be reflected in the learnt representations, and subgroup-level performance requires further study.

In summary, TabText demonstrates that treating tabular health data as language provides a flexible and efficient approach to building predictive models. By capturing contextual information typically discarded in traditional preprocessing, the framework can accelerate model development, support application across institutions, and improve performance in challenging prediction tasks.
\newpage

\subsection*{Data Sharing}\vspace{-5pt}
The patient-level datasets used for patient flow predictions are not publicly available due to the sensitive and confidential nature of hospital data. However, the datasets used in Section~\ref{subsec:PublicDatasets}  are publicly accessible. 
\vspace{-5pt}
\subsection*{Code Availability}\vspace{-5pt}
We provide the complete TabText framework code, along with the data and metadata necessary to replicate our results on the public datasets (\url{https://github.com/kimvc7/TabText} ).
\vspace{-5pt}
\subsection*{Author Contributions}\vspace{-5pt}
K.V.C developed the code, performed the experiments and the corresponding results’ analysis, and co-wrote the paper.
L.N and Y.M developed parts of the code, performed some of the experiments and results’ analysis, and edited the paper. L.B and C.Z performed some of the results’ analysis and edited the paper. L.S participated in the early stages of the research and helped formulate the idea and discussed some of the results. D.B. inspired, supported, and supervised the presented work and edited the paper. All authors take responsibility for the decision to submit for publication.
\vspace{-5pt}
\subsection*{Declaration of Interests}\vspace{-5pt}
All authors declare no financial or non-financial competing interests.

\newpage

\bibliography{main}
\bibliographystyle{unsrtnat}
\newpage 
\appendix
\webextramain{Supplementary Material}
\section{Appendix: Language Construction}\label{sec:language_construction}
In this section, we provide more details about the sentence configurations analysed in Section \ref{sec:language}. We investigate different ways to construct sentences for each column attribute as described below:

\textit{Descriptiveness:} We consider whether or not to use descriptive language to construct text sentences. Specifically, consider a cell from column ``attribute'' that has value ``X''. If the column is non-binary, we consider the following options:
1) {non-descriptive sentence}:  ``attribute: X''; and 2) {descriptive sentence}:   ``attribute is X''.  For binary columns, we consider the verb associated with the specific attribute. For instance, if the column attribute is associated with the verb ``to have''  we consider 1) {non-descriptive sentence}:  ``has X: yes'' or  ``has X: no''; and 2) {descriptive sentence}:   ``has X'' or ``does not have X''.

\textit{Missing Values:} When the value for a column ``attribute'' is missing, we consider two options: to explicitly mention in the text that this information is not available (``attribute is missing''), or to simply skip this column when building the text representation.

\textit{Numerical Data:} 
\textcolor{black}{Discretising continuous features is an effective strategy for deep learning models. \citep{numerical1, gorishniy2021revisiting} Therefore, we also consider whether or not to discretise numerical values by replacing them with text.} For each column, we compute the average (AVG) and standard deviation (SD) on the training data, and then map any value $X$ into one of five categories: ``very low'' if $X < \text{AVG} - 2\text{SD}$; ``low'' if $\text{AVG} - 2\text{SD} \leq X < \text{AVG} - \text{SD}$; ``normal'' if $\text{AVG} - \text{SD} \leq X < \text{AVG} + \text{SD}$; ``high'' if $\text{AVG} + \text{SD} \leq X < \text{AVG} + 2\text{SD}$; and ``very high'' if $X > \text{AVG} + 2\text{SD}$.

\textit{Including Metadata:} We investigate the value of including metadata as part of the text representation. This corresponds to descriptions of table content (e.g., ``This table contains information about the medications administered to this patient'') or the prediction task of interest (e.g., ``We want to predict mortality risk''). 
\section{Appendix: Datasets Information}\label{sec:datasets_info}
This appendix provides supplementary details on the datasets used in our experiments. Table~\ref{table:data_sizes} reports the number of patient-days available for training and testing at hospital HA across all predictive tasks. Table~\ref{tab:summary_tables} summarises the six tabular sources included in the dataset and their associated metadata. Finally, Table~\ref{all_hosp} presents dataset sizes and key characteristics of the six hospitals evaluated in Section~\ref{subsec:rawdata}. Finally, Table~\ref{subsec:Context} describes the metadata used for the public datasets analysed in Section~\ref{subsec:PublicDatasets}.

\begin{table}[ht]
 \caption{Data sizes (number of patient days) for training and testing sets across the nine patient-flow prediction tasks.}
    \centering
    \begin{tabular}{P{5cm}P{3cm}P{3cm}}
\hline\hline
                 \textbf{Prediction Task} &  \textbf{Training}  &\textbf{Testing}  \\
\hline
Discharge 24 hr &  572,964 & 265,917 \\
Discharge 48 hr &  572,964 & 265,917 \\
Enter ICU 24 hr &  385,132 & 180,075 \\
Leave ICU 24 hr &  73,013 &  34,669 \\
Enter ICU 48 hr &  292,659 & 138,947 \\
Leave ICU 48 hr &  68,472 &  33,011 \\
   Expire 24 hr &  572,964 & 265,917 \\
   Expire 48 hr &  572,964 & 265,917 \\
      Mortality &  572,964 & 265,917 \\
\hline\hline
\end{tabular}
\caption{Number of patient days used for hospital HA across tasks.}
    \label{table:data_sizes}
\end{table}

\begin{table}[ht]
    \centering
\begin{tabular}{P{1cm}P{5cm}P{5cm}}
\hline\hline
\textbf{Table \#} & \textbf{Table Meta Information}    & \textbf{Example Columns} \\  \hline       
1        & Lab values                & Platelet, Sodium                            \\
\hline
2        & Chart measurements        & Respiratory rate, oxygen concentration      \\
\hline
3        & Counting statistics       & Number of medications, number of orders     \\
\hline
4        & Current condition         & Oxygen device, is in ICU                    \\
\hline
5        & Historical patient record & Previous admission, previous length of stay \\
\hline
6        & Non-patient-specific data & Day of the week, ward census    \\           \hline\hline
\end{tabular}
\caption{Summary of tabular data, which contains different aspects of a patient’s admission stay from the patient’s high-level demographics to precise lab measurements.}
    \label{tab:summary_tables}
\end{table}

\begin{table}[ht]
\centering
\begin{tabular}{lcccccc}
\hline
\hline
Hospital & Number of Beds & Number of Units & Number of Services & Number of Patient days \\
\hline
HA & 867 & 47 & 48 & 261,954 \\
HB & 233 & 13 & 38 & 52,328 \\
HC & 122 & 11 & 18 & 27,912 \\
HD & 446 & 20 & 34 & 76,325 \\
HE & 156 & 12 & 31 & 39,972 \\
HF & 520 & 26 & 43 & 85,322 \\
\hline
\hline
\end{tabular}
\caption{Hospital metadata (HA–HF): capacity, services, and number of patient-days in 2021.}\label{all_hosp}
\end{table}

\begin{table}[ht]
\centering
\begin{tabular}{P{1cm}P{14cm}}
\hline\hline
\textbf{Dataset} & \textbf{Metadata}  \\ 
\hline
BC & This breast cancer domain was obtained from the University Medical Centre, Institute of Oncology, Ljubljana, Yugoslavia. This is one of three domains provided by the Oncology Institute that have repeatedly appeared in the machine learning literature. (See also lymphography and primary-tumour.) \\
\hline
HD & 4 databases: Cleveland, Hungary, Switzerland, and the VA Long Beach  \\
\hline
TS & The data is dedicated to a classification problem related to the post-operative life expectancy in lung cancer patients: class 1 - death within one year after surgery, class 2 - survival.  \\
\hline
AS & Autistic Spectrum Disorder Screening Data for Adults. This dataset is related to classification and predictive tasks. \\
\hline
CP & Utilise 17 clinical features for predicting the survival state of patients with liver cirrhosis. The survival states include 0 = D (death), 1 = C (censored), 2 = CL (censored due to liver transplantation). \\
\hline
CC & This dataset focuses on the prediction of indicators/diagnosis of cervical cancer. The features cover demographic information, habits, and historical medical records.\\
\hline 
DR & This dataset contains the sign and symptom data of newly diabetic or would-be diabetic patients.\\
\hline 
MM & Discrimination of benign and malignant mammographic masses based on BI-RADS attributes and the patient's age.\\
\hline 
HS & Dataset contains cases from a study conducted on the survival of patients who had undergone surgery for breast cancer.\\
\hline\hline
\end{tabular}
\caption{Metadata information used for datasets from UCI Machine Learning Repository.}
\label{tab:external_datasets}
\end{table}

\begin{table}[ht]
    \centering
    \color{black}
    \begin{tabular}{p{11.6cm}cccccc}
\hline\hline
\textbf{Dataset} & \textbf{Acronym}  & \textbf{n} & \textbf{p} & \textbf{k} \\
\hline
Breast Cancer \citep{breast_cancer_14} & BC & 286 &9 & 2\\
\hline
Heart Disease \citep{heart_disease_45}& HD & 303 & 13 &5\\
\hline
Thoracic Surgery Data \citep{thoracic_surgery_data_277}& TS & 470 & 16 &2\\
\hline
Autism Screening Adult \citep{autism_screening_adult_426}& AS & 704 & 20 &2\\
\hline
Cirrhosis Patient Survival Prediction \citep{cirrhosis_patient_survival_prediction_878}& CP & 418 & 17 &3\\
\hline
Cervical Cancer (Risk Factors) \citep{cervical_cancer_risk_factors_383}& CC & 858 & 36 &2\\
\hline
Early Stage Diabetes Risk Prediction\citep{early_stage_diabetes_risk_prediction_529}& DR & 520 & 16 &2\\
\hline
Mammographic Mass\citep{mammographic_mass_161}& MM & 961 & 5 &2\\
\hline
Haberman's Survival \citep{habermans_survival_43}& HS & 306 & 3 &2\\
\hline
\end{tabular}
\caption{\textcolor{black}{Datasets from the UCI Machine Learning Repository. The columns n, p, and k correspond to data size, data dimension, and number of classes in the prediction task, respectively.}}
\label{tab:datasets-info}
\end{table}

\section{Appendix: Data Preprocessing}\label{sec:data_cleaning}
This section outlines the data preprocessing procedures applied to hospital HA records for the experiments described in Section~\ref{subsec:Context}, as well as metadata information for datasets used in Section \ref{subsec:PublicDatasets}.

\textit{String Parsing:}
Some columns in string format require string parsing to extract numerical features as continuous variables. For instance, the normal ranges of laboratory tests in forms such as ``50--70'' are replaced with two columns: one with a value of 50 for the lower bound and another with a value of 70 for the upper bound. 

\textit{Categorical Encoding:}
Categorical columns (e.g., department,  mobility level, the reason for visit) must be converted to ordered numerical levels (consecutive integers) using label-encoding or binary categories using one-hot encoding. Due to the large number of categories, we use label encoding for all categorical variables. 

\textit{Feature Engineering:}
To better capture the clinical information, we compute various auxiliary variables:
\begin{enumerate}
    \item[1)] Current conditions extracted from records (e.g., whether the patient is in ICU or IV)
    \item[2)] Normal indicators (e.g. whether a measurement is within the normal/critical range) instead of including the ranges themselves.
    \item[3)] Counts (e.g., number of days in ICU, number of attending physicians).
    \item[4)] Pending procedures/results (time until surgery, whether MRI is pending, etc.).
    \item[5)] Historical record linked to the patient (e.g., days since previous admission and length of previous stay).
    \item[6)] Non-patient-specific operational variables (e.g., day of the week, ward census and utilisation, hospital admission volume on the previous day).
\end{enumerate}
\textit{Missing Data Imputation:}
Since the raw data comes from a hospital system, it contains many missing values. We impute most missing entries with 0, except for a few cases.
From communications with the hospital, we impute certain variables with prior knowledge of the meaning of missingness (e.g., missing Do Not Resuscitate (DNR) means the patient did not sign a DNR form). 
For some auxiliary variables, we apply some rules, such as imputing counts with 0 if no record exists and imputing the number of days since previous admission with a large number (e.g., 9999) if no previous admission exists. 
\section{Appendix: Experiment Settings}\label{sec:experiment_details}
This appendix provides additional information on the experimental settings described in Section~\ref{sec:results}. We detail the software implementation, hyperparameter grids, and other relevant implementation considerations. \\

\textcolor{black}{\noindent \textit{Computational Cost:} Compared to the traditional tabular approach, the TabText framework incurs higher computational costs due to fine-tuning, embedding extraction, and training on higher-dimensional data. To keep fine-tuning efficient, we limited all experiments to fewer than 2,000 iterations, ensuring runtimes under 10 minutes. Embedding extraction using Longformer-based LLMs with mean-pooling averaged 0·9 seconds per sample, varying with text length. For our largest dataset (800,000 samples and 160 columns), extraction took approximately 6 hours, while for the smallest datasets (a few hundred samples), it took under 10 minutes. Finally, we observed that when the original tabular data contains only a few dozen columns, adding 768-dimensional embeddings can increase the training time of gradient-boosted trees by up to a factor of 8.}

\noindent \textit{Implementation:} All our code is written in Python 3.8.2. We trained all models using one Intel Xeon Platinum 8260 or Intel Xeon Gold 6248 CPU and GPU. We conducted all of our predictive experiments using the 
scikit-learn~\citep{scikit-learn} 
library from Python. The Clinical-Longformer model is directly accessed from HuggingFace. In Section \ref{subsec:PatFlow}, word-masking fine-tuning for Clinical-Longformer was performed on 2000 randomly sampled training examples (from the training set only, ensuring no overlap with validation or test data) for 3 epochs via word-masking with default hyperparameters. In Section \ref{subsec:PublicDatasets}, word-masking fine-tuning was performed using the entire training set with 7 epochs and a batch size of 4 for all datasets. \\

{\color{black}
\noindent The following are the hyperparameters that we grid-searched to obtain the optimal XGBoost models in Section \ref{subsec:PatFlow}: 
\paragraph{XGBoost.} We searched over:
    \begin{itemize}
        \item Number of estimators: \{100, 200, 300\},
        \item Maximum depth: \{3, 5, 7\},
        \item Learning rate: $\{$0·05, 0·1, 0·3$\}$,
        \item $L_2$ regularisation parameter: \{$1e^{-2}, 1e^{-3}, 1e^{-4}, 1e^{-5}, 0$\}.\\
    \end{itemize}
Regarding Section \ref{subsec:PublicDatasets}, the hyperparameter grids for each model were as follows:
\paragraph{XGBoost.} We searched over:
\begin{itemize}
    \item Number of estimators: $n_{\text{est}} \in \{10, 50, 100, 200\}$
    \item Maximum tree depth: $d_{\max} \in \{3, 5, 7\}$
    \item Learning rate: $\eta \in \{$ 0·01, 0·05, 0·1, 0·5$\}$
    \item $\ell_2$ regularisation strength: $\lambda \in \{0, 10^{-4},$ 0·001, 0·01, 0·1$\}$
\end{itemize}

\paragraph{FTTransformer.} We searched over:
\begin{itemize}
    \item Batch size: $\{8, 16, 32\}$
    \item Learning rate: $\{5\!\times\!10^{-4}, 5\!\times\!10^{-3}\}$
    \item Whether to include continuous features ($\texttt{keep\_cont} \in \{\texttt{True}, \texttt{False}\}$)
    \item Number of training epochs: $\{4, 8, 16\}$
    \item Number of quantile bins used for feature discretisation: $\{5, 10, 50\}$
\end{itemize}

\paragraph{Logistic Regression.} We searched over:
\begin{itemize}
    \item Penalty type: $\in \{\ell_1, \ell_2, \text{elastic net}\}$
    \item Inverse regularisation strength: $C \in \{10^{-5}, 10^{-4}, 10^{-3}, $0·01, 0·1, $1, 10\}$
    \item Elastic net mixing parameter (if applicable): $\alpha \in \{10^{-5}, 10^{-4}, 10^{-3},$ 0·01, 0·1, $1\}$
\end{itemize}}

\end{document}